\documentclass[11pt,a4paper]{article}
\usepackage[hyperref]{acl2021}
\usepackage{times}
\usepackage{latexsym}
\usepackage{graphicx}
\usepackage{multirow}
\usepackage{comment}
\usepackage{subcaption}
\usepackage{longtable}

\newcommand*{\affaddr}[1]{#1} 
\newcommand*{\affmark}[1][*]{\textsuperscript{#1}}
\newcommand*{\email}[1]{\texttt{#1}}

\usepackage{microtype}

\aclfinalcopy 

\title{SemEval-2021 Task 11: \textsc{NLPContributionGraph} -
Structuring Scholarly NLP Contributions for a Research Knowledge Graph}

\author{%
Jennifer D'Souza\affmark[1], S\"oren Auer\affmark[1], and Ted Pedersen\affmark[2]\\ 
\affaddr{\affmark[1]TIB Leibniz Information Centre for Science and Technology, Hannover, Germany}\\
\email{\{jennifer.dsouza, auer\}@tib.eu}\\
\affaddr{\affmark[2] University of Minnesota, Duluth, USA}\\
\email{tpederse@d.umn.edu}%
}

\date{}

\begin{document}
\maketitle
\begin{abstract}

There is currently a gap between the natural language expression of scholarly publications and their structured semantic content modeling to enable intelligent content search. With the volume of research growing exponentially every year, a search feature operating over semantically structured content is compelling. The SemEval-2021 Shared Task \textsc{NLPContributionGraph} (a.k.a. `the NCG task') tasks participants to develop automated systems that structure contributions from NLP scholarly articles in the English language. Being the first-of-its-kind in the SemEval series, the task released structured data from NLP scholarly articles at three levels of information granularity, i.e. at sentence-level, phrase-level, and phrases organized as triples toward Knowledge Graph (KG) building. The sentence-level annotations comprised the few sentences about the article's contribution. The phrase-level annotations were scientific term and predicate phrases from the contribution sentences. Finally, the triples constituted the research overview KG. For the Shared Task, participating systems were then expected to automatically classify contribution sentences, extract scientific terms and relations from the sentences, and organize them as KG triples.  

Overall, the task drew a strong participation demographic of seven teams and 27 participants. The best end-to-end task system classified contribution sentences at 57.27\% F1, phrases at 46.41\% F1, and triples at 22.28\% F1. While the absolute performance to generate triples remains low, in the conclusion of this article, the difficulty of producing such data and as a consequence of modeling it is highlighted.

\end{abstract}

\section{Introduction}

Traditional search models over scholarly communication are now changing toward Knowledge Graph (KG) models operating on structured fine-grained scholarly content offering enhanced contextual search results. Several initiatives exist to this end: Google Scholar, Web of Science~\cite{birkle2020web}, Microsoft Academic Graph~\cite{wang2020microsoft}, OpenAIRE Research Graph~\cite{manghi_paolo_2019_3516918}, Open Research Knowledge Graph~\cite{auer_soren_2018}, Semantic Scholar~\cite{fricke2018semantic} to name just a few. These KG models differ in their content, their level of detail, etc., as they represent diverse aspects of scholarly communication. 

Text, of course, is of seminal importance to Science. It is as important as experimentation itself; unpublished research lacks validity. Seen in another angle, it is hard to imagine a medium other than discourse that can convey a comprehensive picture of the scholarly investigation. For the wider research audience, it is interesting to read the full ``stories'' of Science. 

Nonetheless, since scientific literature is growing at a rapid rate~\cite{johnson2018stm} and researchers today are faced with this publications deluge~\cite{landhuis2016scientific}, it is increasingly tedious, if not practically impossible to keep up with the research progress even within one's own narrow discipline. In this regard, among the existing scholarly knowledge structuring initiatives, the Open Research Knowledge Graph (ORKG)~\cite{auer2020improving} is posited as a solution to the problem of keeping track of research progress minus the cognitive overload that reading dozens of full papers impose. It aims to build a comprehensive KG that publishes the research contributions of scholarly publications per paper, where the contributions are interconnected via the graph even across papers. The ORKG digital library (DL) framework can be accessed here \url{https://www.orkg.org}.

Motivated by the availability of a next-generation DL, we present the SemEval-2021 \textsc{NLPContributionGraph} (NCG) Shared Task as a step in the easier knowledge acquisition of contributions for researchers - \textit{the automated structuring of the unstructured article contributions}. To this end, via the NCG task, we have formalized the building of such a scholarly contributions-focused graph over NLP scholarly articles as an automated task. In the subsequent paper, we detail our task in terms of its resources, organization, participants, and evaluations. 



\section{Data}

The NCG Shared Task comprised a dataset of NLP scholarly articles annotated for their contributions. The contributions were structured to be integrable within KG infrastructures such as the ORKG~\cite{jaradeh2019open} that capture research overviews. The contributions were annotated in three different information granularities, i.e. (1) \textit{Contribution sentences}: a set of sentences about the article's contribution; (2) \textit{Scientific terms and relations}: a set of terms and relational predicates in the contribution sentences; and (3) \textit{Triples}: semantic statements that pair the terms with a predicate, modeled toward subject-predicate-object RDF statements for KG building. This latter set of annotations formed the actual graph. Inspired after article sections, the Triples were organized under three (mandatory) or more of 12 total information units (IUs), viz. \textsc{ResearchProblem}, \textsc{Approach}, \textsc{Model}, \textsc{Code}, \textsc{Dataset}, \textsc{ExperimentalSetup}, \textsc{Hyperparameters}, \textsc{Baselines}, \textsc{Results}, \textsc{Tasks}, \textsc{Experiments}, and \textsc{AblationAnalysis}. 

\subsection{Data Annotation Scheme}

A trial annotation stage preceded the annotation of the Shared Task dataset. In this stage, an annotation scheme was prescribed. This involved specifying the annotation data granularities and the 12 IUs for organizing the triples. Observations were also obtained about the position in the articles where the authors generally stated the contribution. The trial annotations were conducted in two steps: a pilot annotation step~\cite{ncg-pilot-scheme} followed by an adjudication step~\cite{d2020graphing}. The resulting scheme itself was called the \textsc{NLPContributionGraph} (NCG) scheme. 

For the trial stage, a relatively small dataset of 50 articles uniformly distributed across five NLP tasks, i.e. machine translation, named entity recognition, question answering, relation classification, and text classification, were selected. 

Overall, after the pilot annotation task the following core question was answered. Could a scheme be defined such that it would encompass all annotation decisions of the task? In reality, it was found that the scheme could only define high-level annotation decisions such as: where in the article could the contribution information generally be found? E.g., the title, the abstract, a few lines in the Introduction, the first few lines of the Results section. This still entailed making subjective decisions such as if the model is not described in the Introduction then the first few lines of the model description section would need to be annotated. The scheme also specified the 12 IUs for organizing the structured triples. The choice of the specific IU for organizing the triples was based on the closest section title.


After the two-step trial annotation stage, the intra-annotation agreement between the pilot and adjudication steps, in terms of F1, was 67.92\% for sentences, 41.82\% for phrases, and 22.31\% for triple statements indicating that with increased granularity of the information, the annotation adjudication was greater~\citeyearpar{d2020graphing}.

The trial annotations were made by a postdoctoral researcher in Computational Linguistics. The same experienced annotator also annotated the full dataset. Next, we explain the NCG data with a focus on the KG and then offer two supporting examples as illustrations of the data. 

\subsection{Understanding our Knowledge Graph}

The \textsc{NCG} KG used two levels of knowledge systematization: 1) At the root, it defined a dummy node called \textsc{Contribution}. And following the root node, 2) it defined the 12 nodes introduced earlier and generically referred to as Information Units or IUs. Each scholarly article's annotated contribution triple statements were organized under three (mandatory) or more of these IU nodes, depending on whether they applied to the article. Next, we provide details about each IU. 

\paragraph{\textsc{ResearchProblem}} The research challenge addressed by a contribution. In other words, a focus of the research investigation or the issue for which a research solution was proposed.

\paragraph{\textsc{Approach} or \textsc{Model}} The contribution of the paper as the solution proposed for the research problem. This unit was called \textsc{Approach} when the solution was proposed as an abstraction, and was called \textsc{Model} if the solution was proposed in practical implementation terms. Further, in case the solution was not referred to as approach or model in the article, the reference was normalized as either \textsc{Approach} or \textsc{Model}. E.g., references like ``method'' or ``application'' were normalized as \textsc{Approach}; on the other hand, references like ``system'' or ``architecture,'' were normalized to \textsc{Model}. This unit captured only proposed system highlights.

\paragraph{\textsc{Code}} The contribution resource; the link to the software on an open-source hosting platform such as Gitlab or Github or on the author's website.

\paragraph{\textsc{Dataset}} Like \textsc{Code}, this a contributed resource in the form of a dataset.

\paragraph{\textsc{ExperimentalSetup} or \textsc{Hyperparameters}} Details about the platform including both hardware (e.g., GPU) and software (e.g., Tensorflow library) for implementing the machine learning solution; and of variables, that determine the network structure (e.g., number of hidden units) and how the network is trained (e.g., learning rate), for tuning the software to the task objective. It was called \textsc{ExperimentalSetup} only when hardware details were provided.

\paragraph{\textsc{Baselines}} The systems that a proposed \textsc{Approach} or \textsc{Model} were compared with.

\paragraph{\textsc{Results}} The main findings or outcomes reported in an article for the \textsc{ResearchProblem}.

\paragraph{\textsc{Tasks}} The \textsc{Approach} or \textsc{Model}, particularly in multi-task settings, are tested on more than one task, in which case, this unit was defined to capture all the experimental tasks. Unlike the earlier units, the \textsc{Tasks} IU was a container for more than one of the earlier mentioned IUs. Specifically, each task listed in \textsc{Tasks} could include one or more of the \textsc{ExperimentalSetup}, \textsc{Hyperparameters}, and \textsc{Results} as sub-information units.

Furthermore, since it is common in NLP for tasks to be defined over datasets, experimental tasks are often synonymous with the experimental datasets, therefore this unit was also applied in articles where the datasets were explicitly listed instead of the task names.

\paragraph{\textsc{Experiments}} The second container information unit, like \textsc{Tasks}, defined to include one or more of the previous discussed units as sub-information units. This unit encapsulated several \textsc{Tasks} themselves and consequently, the units that \textsc{Tasks} encapsulated, i.e. \textsc{ExperimentalSetup} and \textsc{Results}, or a combination of \textsc{Approach}, \textsc{ExperimentalSetup} and \textsc{Results}.

\paragraph{\textsc{AblationAnalysis}} A form of \textsc{Results} that describes the performance of components in an \textsc{Approach} or \textsc{Model}. 

\subsection{Data Examples}

Below, we show two examples of two different IUs, viz. \textsc{ResearchProblem} and \textsc{Model}, respectively, as illustrations of our data.

\begin{figure}[!htb]
\includegraphics[width=\linewidth]{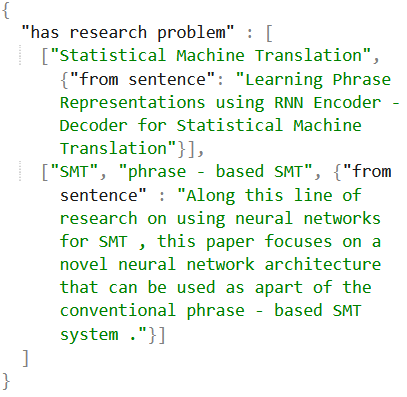}
\caption{Annotated data in JSON format for the \textsc{ResearchProblem} Information Unit for the paper ``Learning Phrase Representations using RNN Encoder–Decoder for Statistical Machine Translation.''}
\label{research-problem-json}
\end{figure}

\paragraph{Example 1}

In this example, the \textsc{ResearchProblem} IU is modeled for the following reference paper: \textit{Learning Phrase Representations using RNN Encoder–Decoder for Statistical Machine Translation}~\cite{cho2014learning}. We show two formats of our data: the JSON format (see Fig.~\ref{research-problem-json}) with all three annotated information granularities; and the triples format (see Table~\ref{research-problem-triples}) showing only the annotated data for a KG. In the JSON data, the dummy root node \textsc{Contribution} is left unspecified, however, it is specified in the triples. For this data, three phrases that named the research problem were annotated. The phrases were attached to the dummy root node by the predicate ``has research problem.'' Further, in the JSON data, following the predicate ``from sentence,'' the selected contribution sentences are listed.

\begin{table}[!htb]
\small
\begin{tabular}{|p{7cm}|} \hline
(Contribution, has, Statistical Machine Translation) \\
(Contribution, has, SMT) \\ 
(Contribution, has, Phrase - Based SMT) \\ \hline
\end{tabular}
\caption{Annotated \textsc{ResearchProblem} Information Unit contribution data as triples. This data is obtained from the JSON data shown in Fig~\ref{research-problem-json}.}
\label{research-problem-triples}
\end{table}

\begin{figure}[!htb]
\includegraphics[width=\linewidth]{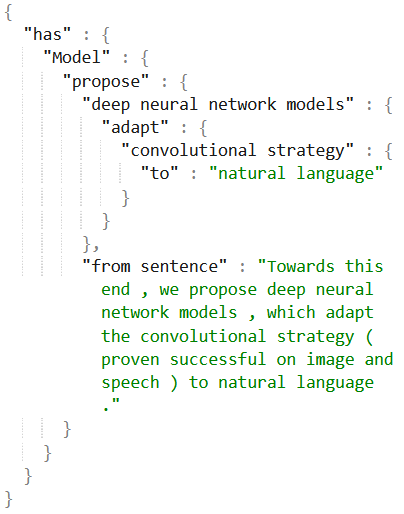}
\caption{Annotated data in JSON format for the \textsc{Model} Information Unit for the paper ``Convolutional Neural Network Architectures for Matching Natural Language Sentences.''}
\label{model-json}
\end{figure}

\paragraph{Example 2} In this example, a subpart of the \textsc{Model} IU is annotated for the following reference paper: \textit{Convolutional Neural Network Architectures for Matching Natural Language Sentences}~\cite{hu2014convolutional}. See Fig.~\ref{model-json} for the JSON format and Table~\ref{model-triples} for the triples data.

\begin{table}[!htb]
\small
\begin{tabular}{|p{7cm}|} \hline
(Contribution, has, Statistical Machine Translation) \\
(Model, propose, deep neural network models) \\
(deep neural network models, adapt, convolutional strategy) \\
(convolutional strategy, to, natural language) \\ \hline
\end{tabular}
\caption{Annotated \textsc{Model} Information Unit contribution data as triples. This data is obtained from the JSON data illustrated in Fig.~\ref{model-json}.}
\label{model-triples}
\end{table}

\begin{table*}[!htb]
\small
\centering
\begin{tabular}{|l|p{0.8cm}|l|l|p{1.4cm}|p{1.4cm}|l|l|l|} \hline
Dataset & \textit{info-units} & \textit{sentences} & \textit{entities} & \textit{total triples} & \textit{total unique triples} & \textit{subject} & \textit{predicate} & \textit{object} \\ \hline
\textsc{Trial}   & 217 & 1,029 & 4,777 & 2,924 & 2,782 & 1,427 & 1,181 & 2,512   \\
\textsc{Train}   & 1,050 & 5,064 & 30,485 & 18,679 & 17,356 & 8,173 & 4,538 & 13,335 \\
\textsc{Test} & 642 & 2,720 & 16,435 & 10,623 & 10,002 & 4,951 & 2,447 & 8,282 \\ \hline
\end{tabular}
\caption{\textsc{NlpContributionGraph} Shared Task 2021 Overall Corpus Statistics}
\end{table*}

\subsection{Data Statistics}


Overall, the \textsc{NCG} Shared Task dataset had 50 articles in the trial data, 237 articles in the training data, and 155 articles in the test data. The trial data articles uniformly spanned five tasks, the training data spanned 24 tasks, and the test data spanned 10 tasks. For the Shared Task itself, participants were encouraged to merge the trial and training datasets. Thus, the overall training data had 287 articles representing 29 unique tasks. The training and test tasks were mutually exclusive except for one, i.e. `natural language inference.' Table 3 shows further detailed statistics of the \textsc{NCG} dataset in terms of each of the annotated information granularities.

Our full dataset is publicly released online~\cite{ncg-final-dataset}.

\section{Task Description}

Our comprehensive \textsc{NCG} Shared Task formalism was as follows. Given a scholarly article $A$ in plaintext format, the goal was to extract (1) a set of contribution sentences $C_{sent} = \{C_{sent_{1}}, ... ,C_{sent_{N}}\}$, (2) a set of scientific knowledge terms and predicates from $C_{sent}$ referred to as entities $E = \{e_{1}, ... ,e_{N}\}$, and (3) to organize the entities $E$ as a set of (subject,predicate,object) triple statements $T = \{t_{1}, ... ,t_{N}\}$ toward KG building organized under three or more of the 12 total IUs.

\paragraph{Task Evaluation Phases.} The task comprised three evaluation phases, thereby enabling detailed system evaluations.

\noindent{\textit{Evaluation Phase 1: End-to-end Pipeline.}} In this phase, systems were tested for the comprehensive end-to-end KG building task described in the formalism above. Given a test set of articles $A$ in plaintext format, the participating systems were expected to return: (1) a set of contribution sentences $C_{sent}$, (2) a set of scientific knowledge terms and predicates from $C_{sent}$, i.e. entities $E$, and (3) the entities in $E$ organized in a set of triple statements $T$ toward KG building. System outputs were evaluated for the three aspects and overall.

\noindent{\textit{Evaluation Phase 2, Part 1: Phrases and Triples.}} In this phase, systems were tested only for their capacity to extract phrases and organize them as triples. Given a test set of articles $A$ in plain-text format and contribution sentences $C_{sent}$ from each article, each system was expected to return: (1) the entities $E$, and (2) the set of triple statements $T$. 

\noindent{\textit{Evaluation Phase 2, Part 2: Triples.}} In this phase, systems were tested only for the triples formation task. Thus, given gold entities $E$ for the set of $C_{sent}$, systems were expected to form triple statements $T$.

\section{Task Setup}

\paragraph{Online Competition}

We used the CodaLab platform for running the competition online \url{https://competitions.codalab.org/competitions/25680}. For the convenience of the participants, the task was divided into four phases. In the Practice phase, which began on Aug 16, 2020, we
released the participant kit that included the full training dataset along with the Python code of the official scoring program \url{https://github.com/ncg-task/scoring-program}. In the Evaluation phases that lasted from Jan 10 till Feb 1, 2021, we provided the participants with masked versions of the test set based on the current evaluation phase. The test set annotations in each phase were uploaded to CodaLab and were not available to the participants. To obtain results, the participants were expected to upload their system outputs to Codalab where they were automatically evaluated by our script and reference data stored on the platform. In each evaluation phase, teams were restricted to make only 10 submissions and only one result, i.e. the top-scoring result, was shown on the leaderboard.

Before the task began, our participants were onboarded via our task website \url{https://ncg-task.github.io/}. Further, participants were encouraged to discuss their task-related questions via our task Google groups page at \url{https://groups.google.com/forum/#!forum/ncg-task-semeval-2021}.

\paragraph{The \textsc{NCG} Data Collection of Articles}

Our base collection of scholarly articles was downloaded from the publicly available leaderboard of tasks in AI called \url{https://paperswithcode.com/}. While \textit{paperswithcode} predominantly represents the NLP and Computer Vision research fields in AI, we restricted ourselves just to its NLP papers. From their overall collection of articles, the tasks and articles in our final data were randomly selected. The raw articles' pdfs needed to undergo a two-step preprocessing before the annotation task. 1) For pdf-to-text conversion, the GROBID  parser~\cite{GROBID} was applied; following which, 2) for plaintext pre-processing in terms of tokenization and sentence splitting, the Stanza toolkit~\cite{qi2020stanza} was used. The resulting pre-processed articles could then be annotated in plaintext format. Note, our data consists of articles in English.


\paragraph{Evaluation Metrics}

The \textsc{NCG} Task participating team systems were evaluated for classifying contribution sentences, extracting scientific terms and relations, and extracting triples (see specific details in Section 3). The results from the three evaluations parts were also cumulatively averaged as a single score to rank the teams. Finally, for the evaluations, the standard precision, recall, and F1-score metrics were leveraged. 

This completes our discussion of the \textsc{NCG} task in terms of its dataset definition and overall organization description. In the remainder of the paper, we shift our focus to the participating teams. Specifically, we describe the participating systems and examine their results for the \textsc{NCG} task.

\section{Participating System Descriptions}

The \textsc{NCG} Shared Task received public entries from 7 participating teams in all. In this section, we briefly describe the teams' systems in terms of the three parts of the \textsc{NCG} task, i.e. contribution sentence classification, scientific terms and relations extraction, and triples extraction.

\subsection{Contribution Sentence Classification} 

To identify the contribution sentences from articles, systems adopted one of two strategies: a binary classification objective, or a multi-class classification objective. In the first strategy, sentences were either classified as contribution sentences or not. In the second strategy, sentences were classified in a 13-class classification task as one of the 12 IUs or as a non-contribution sentence. Next, we describe these strategies. Note, the asterisk superscripts against team names, where present, correspond to ${***}$ 3rd best, ${**}$ 2nd best, and ${*}$ 1st best systems in the Shared Task, respectively.

\paragraph{Binary Classifiers} Team YNU-HPCC~\cite{ynu-hpcc} employed BERT as a \textit{binary classifier} to classify the contribution sentences. Team INNOVATORS~\cite{innovators} also employed a BERT-based \textit{binary classifier} wherein each instance was a set of 10 sentences with additional sentences as context features to the model. Team KnowGraph@IITK$^{***}$~\cite{knowgraph-iitk} used the standard SciBERT + BiLSTM architecture~\cite{Beltagy2019SciBERT} as a \textit{binary sentence classifier}. Team UIUC\_BioNLP$^{*}$~\cite{uiuc-bionlp} employed BERT-based \textit{binary sentence classifier} with features that handled sentence characteristics w.r.t. their context in the article - specifically, its closest preceding topmost and innermost section headers and its position in the article.

\paragraph{Multi-class Classifiers} Team DULUTH~\cite{duluth} framed a \textit{13-class multi-class classification} task. They employed deBERTa~\cite{he2020deberta} as their classifier. Team ECNUICA~\cite{ecnuica} employed three pre-trained transformer models, viz. RoBERTa~\cite{liu2019roberta}, SciBERT~\cite{Beltagy2019SciBERT}, and BERT~\cite{devlin2019bert} as an ensemble classifier. They formulated a multi-class classification task as well. The features to BERT models are the original sentence, contextual information as previous and next sentence to the original sentence, and a sub-title of the paragraph with the separator token ([SEP]) in between. Team ITNLP$^{**}$~\cite{itnlp} employed a BERT-based multi-class classifier that leveraged sentence context and the paragraph heading as additional features.

These binary and multi-class sentence classifiers, were also adapted to our following dataset characteristics.

\subsubsection{Contribution sentences data imbalance} Characteristically, of all the sentences in training data scholarly articles, only 10\% were annotated as contribution sentences. Thus, our dataset presented an imbalanced classification task. 

Teams YNU-HPCC, DULUTH, KnowGraph@IITK$^{***}$ and UIUC\_BioNLP$^{*}$ trained their classifiers on the given data. While INNOVATORS and ITNLP$^{**}$ downsampled the non-contribution sentences. INNOVATORS established a threshold based on cumulative contributing sentence bigram scores as a filter; ITNLP fixed the ratio of positive to negative samples as an integer and tuned the value.

\subsubsection{Differing tasks coverage between the training and the test datasets}

Since only one task was in common between the training and the test datasets, this meant that systems trained only on the training data would be applied on articles from nine new tasks as test data. To this end, Team ECNUICA hypothesized that if the classifier could \textit{see}, i.e. somehow be trained on, the test data tasks, its performance could be boosted. They, thus, adopted the strategy of retraining their classification ensemble with silver-labeled test data instances. This followed the standard setup of training the classifier on the actual training data, applying it to the test data, and incrementally retraining the classifier leveraging the few confidently classified test instances. The instances were marked as silver training data only when all three ensemble classifiers predicted the same class.

\subsection{Scientific Terms and Relations Extraction}

After identifying the contribution sentences, systems then had to extract their scientific terms and relational predicates.

\paragraph{Sequence Labeling Systems} Majority, i.e. six, of the seven participating systems adopted a sequence labeling approach.

\begin{enumerate}
\item Team YNU-HPCC used a pre-trained BERT model for sequence labeling of each token, obtaining embeddings for each token in the sequence, with softmax and argmax top layers which were shared across all tokens.

\item Team DULUTH trained a feature-based maximum-entropy Markov model (MEMM) to predict scientific terms in the contribution sentences.

\item Team ECNUICA extracted entities using RoBERTa~\cite{liu2019roberta} with a CRF layer and a BIO sequence labeling scheme. The input sequences to RoBERTa are modified with sub-title information.

\item Team KnowGraph@IITK$^{***}$ extracted phrases in the sentence by adding BiLSTM layers to the SciBERT + CRF model as a sequence labeler. To mark phrase boundaries, they used the BILUO scheme.

\item Team ITNLP$^{**}$ employed the standard BERT-based model, however, in a sequence labeling setting. They trained ten different models by 10-fold cross-validation and used a voting count threshold scheme to extract the final set of entities.

\item Team UIUC\_BioNLP$^{*}$ used a BERT-CRF model
for phrase extraction and type classification~\cite{souza2019portuguese}. They employed the BIO scheme to distinguish the scientific terms vs. predicate phrases. 

\end{enumerate}

\paragraph{Rule-based System} Team INNOVATORS leveraged an unsupervised rule-based approach for phrase extraction. Using spaCy~\cite{spacy}, they obtained dependency parses for each sentence. They then implemented a set of dependency tree node traversal heuristics for phrase extraction based on the dependency parses. 

\begin{table*}[!htb]
\small
\begin{tabular}{p{11.8cm}|p{0.8cm}|p{0.8cm}|p{0.8cm}}
 & 1 & 2.1 & 2.2 \\ \hline
\textbf{ITNLP} \url{https://github.com/itnlp606/nlpcb-graph} & \textbf{47.03} & 68.63 & 79.31 \\ \hline
\textbf{UIUC\_BioNLP} \url{https://github.com/Liu-Hy/nlp-contrib-graph} & 38.28$^{**}$ & \textbf{76.12}  & \textbf{85.94} \\ \hline
\textbf{KnowGraph@IITK} \url{https://github.com/sshailabh/SemEval-2021-Task-11} & 37.83 & 63.18 & 76.0 \\ \hline
\textbf{ECNUICA} & 33.35 & 71.13 & 81.45 \\ \hline
\textbf{INNOVATORS} \url{https://github.com/HardikArora17/SemEval-2021-INNOVATORS} & 32.05 & 52.52 & 59.71 \\ \hline
\textbf{DULUTH} \url{https://github.com/anmartin94/DuluthSemEval2021Task11} & 28.38 & 49.21 & 45.62 \\ \hline
\textbf{YNU-HPCC} \url{https://github.com/maxinge8698/SemEval2021-Task11}  &  & 75.79 & 65.41
\end{tabular}
\caption{The seven \textsc{NLPContributionGraph} participating teams with their averaged F1 scores over individual subtasks per evaluation phase. \small{Column ``1'' - \textit{Evaluation Phase 1: End-to-end Pipeline} F1; Column ``2.1'' - \textit{Evaluation Phase 2, Part 1: Phrases and Triples} F1; and Column ``2.2'' - \textit{Evaluation Phase 2, Part 2: Triples Extraction} F1.} \\ \footnotesize{**system submission had error in phrase offsets for task submission; actual task performance was 49.72 F1.}}
\end{table*}

\subsection{Triples Extraction}

\begin{enumerate}
\item Team YNU-HPCC first classified the scientific terms in subject, predicate, and object roles using three binary BERT classifiers. These triples from each contribution sentence were then organized as the 12 IUs leveraging a 12-class \textit{contribution} sentence classifier. This team, however, did not participate in the end-to-end evaluation task.

\item Team DULUTH applied Stanford Core NLP’s dependency parser~\cite{chen2014fast} to generate a dependency parse for each contribution sentence. They used the dependency parse structures to assign subject, relation, and object phrase roles to the extracted scientific terms. These were then organized as triples per IU obtained by their 13-class sentence classifier. \textit{The overall end-to-end pipeline system score achieved by this system is 28.38\%.}

\item Team INNOVATORS implemented a set of rules based on the dependency parses to form triples from the extracted scientific terms. They used a CNN-based architecture for classifying the \textit{contribution} sentences as the 12 IUs. \textit{Their end-to-end score was 32.05\%.}

\item Team ECNUICA approached the triples formation task in two steps: i) they formed triple candidates based on the scientific term sequence order in the sentence. Additionally, they employed a set of predefined predicates when the predicates were not directly found in the sentence. ii) They then employed a SciBERT-based binary classifier to classify the triples as true or false candidates. \textit{Their overall end-to-end system score was 33.35\%.}

\item Team KnowGraph@IITK$^{***}$ addressed the \textsc{ResearchProblem}, \textsc{Code}, \textsc{Baselines} and \textsc{AblationAnalysis} IUs by a heuristics-based approach. For the remaining eight IUs triples, they followed a 3-step approach: i) identify predicates from the scientific terms using a binary SciBERT+BiLSTM classifier; and ii) formed triples by arranging the terms and predicates in exact order as they appear in the original sentence; and iii) employ an 8-class SciBERT + BiLSTM classifier to classify the triples. \textit{Their overall end-to-end system score was 37.83\%.}

\item Team ITNLP$^{**}$ extracted triples as follows: i) they formed all possible triples candidates from the classified scientific terms; and ii) employed a binary BERT classifier for true or false candidates. Prior to BERT classification, they perform the negative candidate triples downsampling as follows: by artificially generating them using random replacement (RR) of one of the arguments of the true triples with a false argument; and by random selection (RS) of triples where no argument is a valid pair of another. Additionally, each of their system components obtained boosted performances with the Friendly Adversarial Training strategy~\cite{zhang2020attacks}. \textit{Their overall end-to-end system score was 47.03\%.}

\item Team UIUC\_BioNLP$^{*}$ categorized the triples into six types based on our dataset characteristics. Four of the six types were: structuring intra-sentence information; linking sentence information to IU; linking IU to the root node; and structuring inter-sentence information. The first two of the four broad types were further subdivided into two based on whether the predicate was found in the sentence or was the term ``has.'' Each of the six types were addressed by a specifically trained BERT classifier. \textit{They obtained an overall end-to-end system score of 38.28\% within the task deadline and 49.72\% a day later after fixing phrase component offset errors.}\footnote{Per the task timeline, i.e. within the Phase 1 end-to-end system evaluation, the team achieved 38.28\% F1 within the task deadline due to an error in their submission offsets for phrases. Thus, they are officially 2nd after the ITNLP team within the Shared Task timeline for Phase 1.}

\end{enumerate}

\begin{table*}[!tb]
\small
\resizebox{\textwidth}{!}{\begin{tabular}{|l|c|l|l|c|l|l|l|l|l|c|l|l|}
\hline
\multicolumn{1}{|c|}{\multirow{2}{*}{Model}} & \multicolumn{3}{c|}{Sentences}          & \multicolumn{3}{c|}{Phrases}            & \multicolumn{3}{l|}{Information Units} & \multicolumn{3}{c|}{Triples}            \\ \cline{2-13} 
\multicolumn{1}{|c|}{} & \multicolumn{1}{l|}{F1} & P & R & \multicolumn{1}{l|}{F1} & P & R & F1 & P & R & \multicolumn{1}{l|}{F1} & P & R     \\ \hline
UIUC\_BioNLP    & 57.27 & \textbf{53.61} & 61.46 & 46.41 & \textbf{42.69} & 50.83 & 72.93 & 66.67 & 80.49 & 22.28 & \textbf{22.3}  & \textbf{22.26} \\ \hline
ITNLP           & 56.19 & 51.74 & 61.46 & 45.22 & 41.6  & 49.55 & 72.93 & 66.67 & 80.49 & 13.79 & 13.39 & 14.23 \\ \hline
KnowGraph@IITK  & 46.8  & 39.69 & 57.01 & 35.4  & 28.99 & 45.44 & 60.54 & 44.13 & \textbf{96.34} & 8.57  & 6.53  & 12.45 \\ \hline
ECNUICA         & 39.78 & 26.21 & \textbf{82.48} & 32.03 & 20.73 & \textbf{70.37} & 54.05 & 42.86 & 73.17 & 6.78  & 4.28  & 16.29  \\ \hline
INNOVATORS      & 39.87 & 39.32 & 40.45 & 15.63 & 13.27 & 19.01 & 71.72 & \textbf{82.54} & 63.41 & 0.97  & 14.29 & 0.5   \\ \hline
DULUTH          & 38.1  & 44.83 & 33.12 & 7.08  & 13.07 & 4.86  & 64.41 & 60.0  & 69.51 & 3.94  & 9.2   & 2.51  \\ \hline
\end{tabular}}
\caption{\textit{Evaluation Phase 1: End-to-end Pipeline} Results}
\end{table*}

\begin{figure*}[!htb]
\begin{subfigure}{.48\textwidth}
  \centering
  \includegraphics[width=.8\linewidth]{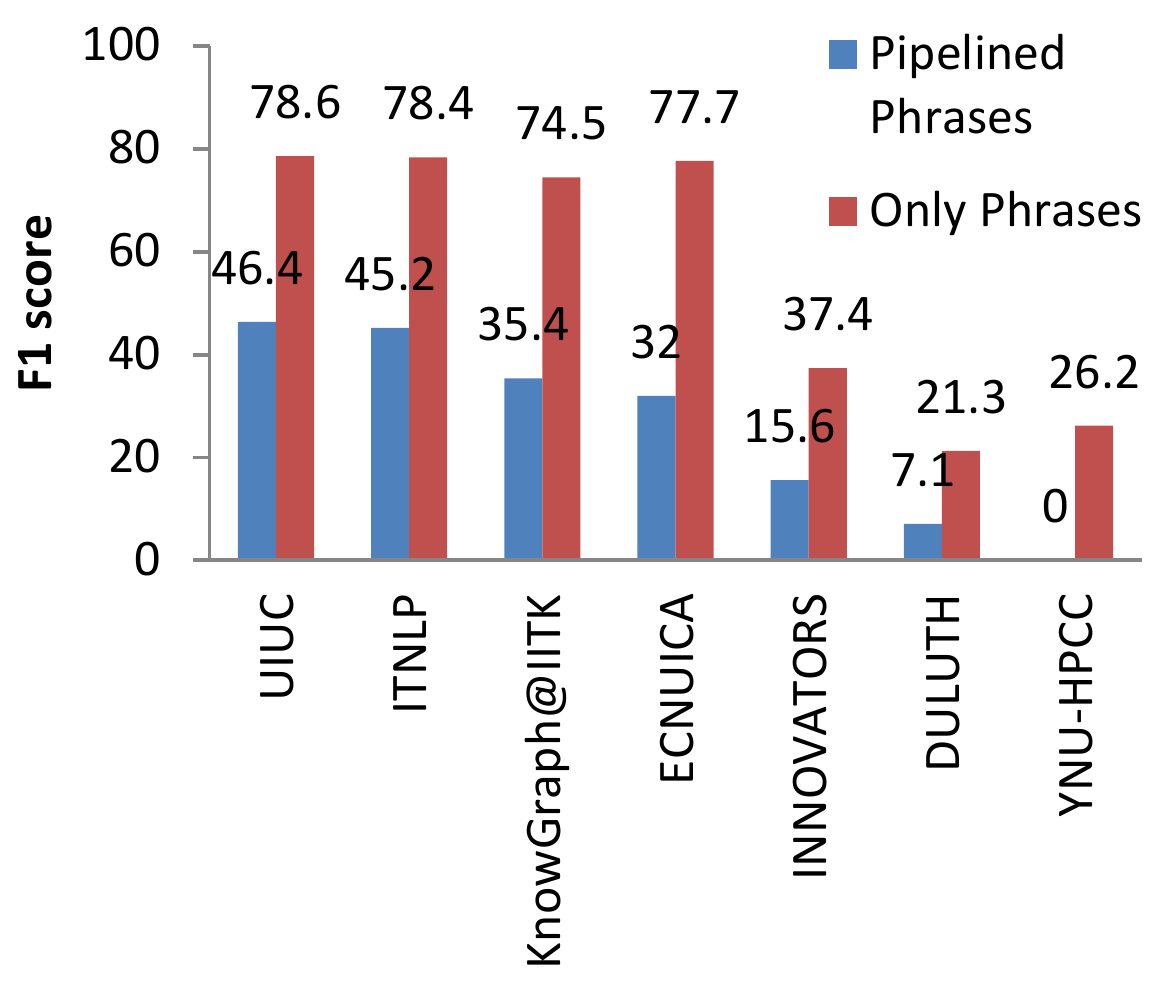}
  \caption{\textit{Evaluation Phase 1: End-to-end Pipeline} \underline{Pipelined Phrases} (blue bars) and \textit{Evaluation Phase 2, Part 1: Phrases and Triples} \underline{Only Phrases} (red bars) results}
  \label{fig:sub-first}
\end{subfigure}
\begin{subfigure}{.51\textwidth}
  \centering
  \includegraphics[width=\linewidth]{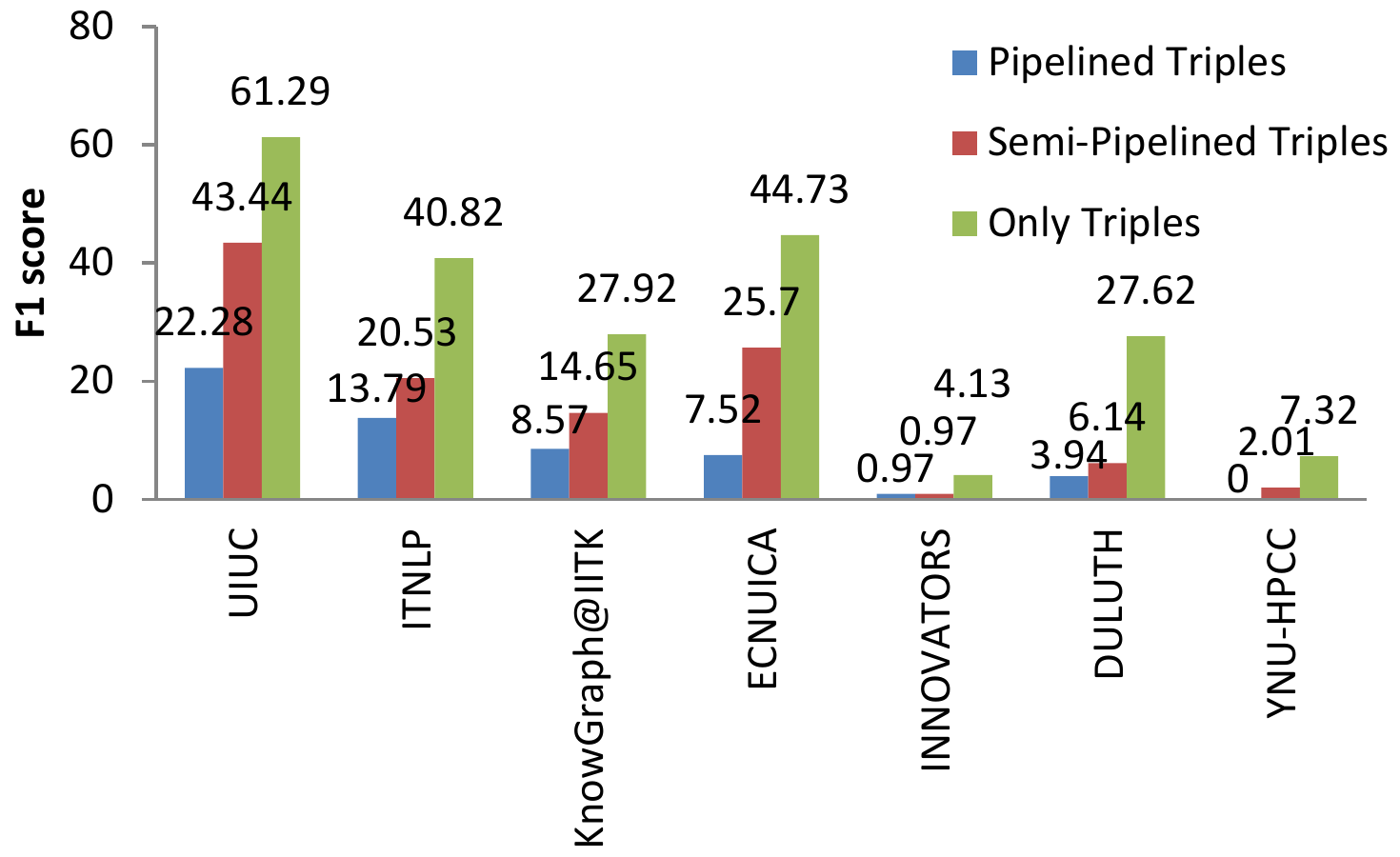}
  \caption{\textit{Evaluation Phase 1: End-to-end Pipeline} \underline{Pipelined Triples} (blue bars); \textit{Evaluation Phase 2, Part 1: Phrases and Triples} \underline{Semi-Pipelined Triples} (red bars); and \textit{Evaluation Phase 2, Part 2: Triples} \underline{Only Triples} (green bars) results}
  \label{fig:sub-second}
\end{subfigure}
\caption{(a) Phrases and (b) Triples extraction results}
\label{fig:fig}
\end{figure*}

\section{Shared Task Results}

In this section, we present the results of the seven participating teams' systems. 

The results in Table 4 show the cumulative scores of the participating teams in each of the three evaluation phases in our Shared Task. We refer the reader to Section 3 for a detailed description of the three evaluation phases. In each phase, Teams were officially ranked by these scores. Next, we examine the scores by the individual extraction tasks that constituted building the \textsc{NLPContributionGraph} per article.

\subsection{Contribution Sentences Classification}

As a first step toward building the \textsc{NLPContributionGraph}, systems were evaluated for identifying contribution sentences. This was done only in the Evaluation Phase 1 of the Shared Task, i.e. the phase that tested the end-to-end systems. These results are shown in Table 5 under column ``Sentences.'' This subtask attained a high score of 57\%. The top two teams, i.e. UIUC\_BioNLP$^{*}$ and ITNLP$^{**}$, differed by only 1 point. Comparing these performances to a baseline, a default system would return all titles as candidate contribution sentences. This results in a score of 10.78\% F1 at 90\% precision and 5.7\% recall. In contrast to the 1 sentence per article result in the default computation, our actual data averages at 17 sentences per article. Thus the default score was computed on a significantly underestimated data sample as also reflected by its low recall. Nevertheless, the top systems significantly outperform this default score with both systems averaging at 20 sentences per article. The least score was also significantly better than the default at 38.1\% F1 at an average of 12 sentences per article.

With F1 less than 60\%, the task shows itself challenging. Some teams ascribed this to the dataset characteristic that contribution sentences constituted only a minority of the sentences in the article ($<$10\%) and thus, overall, presented imbalanced data. To address this they downsampled the data. However, from the two participant systems that used a downsampling strategy, it could not be conclusively verified as an effective strategy since these systems performed on opposite ends of the performance spectrum. On the other hand, \textit{incorporating the closest section header and sentence position as features in the BERT model showed itself an effective and reliable strategy for sentence classification.} This modeled the dataset better since the sentences were annotated from a few sections and the sentences were usually close to the section header. The system UIUC\_BioNLP$^*$ that incorporated such features outperformed all other systems including the ones with the downsampling strategy, i.e. ITNLP$^{**}$ and INNOVATORS.

Finally, how did bootstrapping the test data as silver-labeled data impact model performance? Team ECNUICA that adopted this strategy did not obtain a balanced harmonic mean between their precision and recall achieving the highest recall among all teams of 82.48\% and the lowest precision of 26.21\%. Thus this strategy did not show itself too effective and reliable.

\subsection{Scientific Terms and Relations Extraction}

These results are shown in Table 5 under column ``Phrases''  for the end-to-end systems. The highest F1 obtained on this task was 46.41\%. However, this score was impacted by the pipeline setup such that the low performance in sentence classification impacted the performance in this stage. 
We conducted a separate evaluation phase to control for this aspect. In other words, we examined how would the systems perform only on extracting terms and relations given gold contribution sentences? These results are shown in Figure 3 (a). In fact, the bar chart offers a perspective on the significant differences in system performances when applied on automatically extracted sentences versus gold data. The systems showed the same performance ranking order in both settings. This is a somewhat expected result since none of the systems implemented any specific noisy sentence handling strategy in which case performance differences may have risen. In conclusion, the best result was 46.4\% F1 in the end-to-end setting and was 78.6\% F1 when given gold sentences. 

Notably, the pipeline systems were 10 points lower for extracting phrases than for sentences.

\subsection{Triples Extraction}

The final extraction task to build the \textsc{NCG} per article was to form triples from the extracted terms and relations. These results for the pipeline systems are shown in Table 5 under column ``Triples.'' 
The best performance was 22.28\% F1 and the 2nd best was significantly lower at 13.79\% F1. To evaluate system performances purely for extracting triples, thereby cancelling out the effect of the pipeline setup, additional evaluations were conducted wherein gold data were incrementally made available to the system. These results are shown in Figure 3 (b). Given only the gold sentences, the best team attained 43.44\% F1; given gold terms and relations in addition, they achieved 61.29\% F1. A score of 61.29\% F1 is a strong performance on a still fairly difficult task given the annotation decision subjectivity that may have crept into the data thereby producing considerable variations in annotation patterns. This is discussed in Section 7.

\paragraph{Identifying only the Information Unit Labels} We conducted a meta-evaluation for identifying the set of IU labels per article. These results are shown in Table 5 under column ``Information Units.'' The top two teams were tied at 72.93\% F1 with the second best score at 60.54\% F1. Like sentence classification, a default system could be implemented for this task as one that output just the three mandatory IUs, i.e. \textsc{ResearchProblem}, \textsc{Model}, and \textsc{Results} for all articles. The scores from this default system were 69.01\% F1, 81.67\% precision, and 59.76\% recall. It is 9 points better than the 2nd best. When given gold sentences, systems could be evaluated for identifying just the IUs since the classification were dependent on the underlying sentences.
These results are shown in Fig. 4.

A notable exception in the results is that the IU classification score by Team INNOVATORS remained unchanged regardless of pipelined or gold sentences as input. This is because their downsampling heuristic once designed did not rely on the underlying data when filtering. It is likely that the new gold sentences information was not used at all.

\begin{figure}[!htb]
  \centering
  \includegraphics[width=\linewidth]{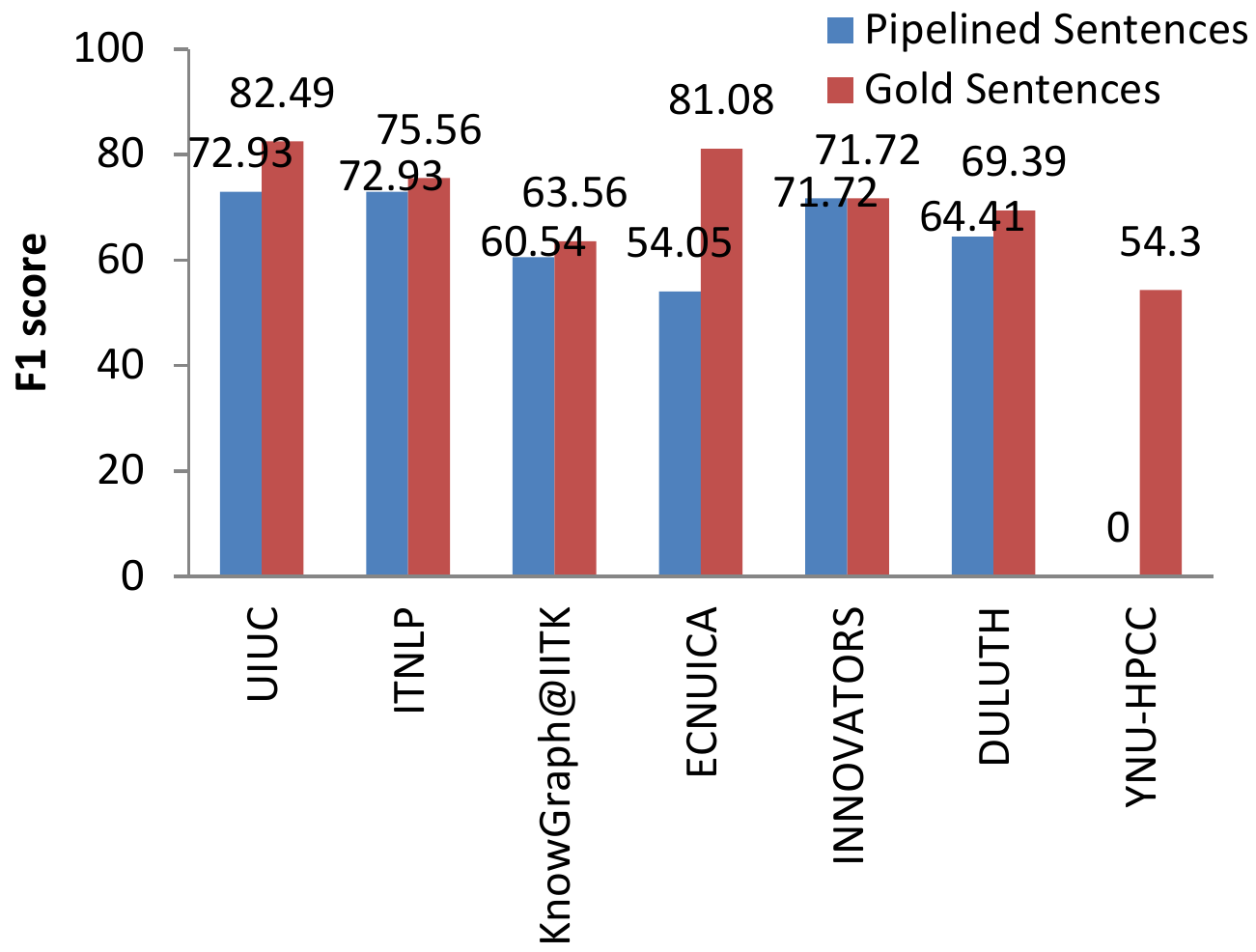}
  \caption{Information Unit identification results in \textit{Evaluation Phase 1: End-to-end Pipeline} with \underline{Pipelined Sentences} (blue bars) and \textit{Evaluation Phase 2, Part 1 and Part 2} with \underline{Gold Sentences} (red bars)}  
\end{figure}

\section{Discussion}

Finally, we conclude our Shared Task paper with a discussion on the perceived limitations of our dataset that can potentially be addressed in future work. Thereby, a new dataset will present new opportunities to evaluate systems on this novel task.

\paragraph{Single Annotator Annotations} The \textsc{NCG} Shared Task dataset was annotated by a single annotator. Further, the design of the annotation scheme was supported by only an intra-annotator consensus agreement score for that annotator. Since this work is the first-of-its-kind in proposing an initial scheme, and given the complex nature of this annotation task with the need to design a model within a realistic timeframe, our annotation procedure is well-suited. 

However, as discussed in our related work~\cite{d2020graphing}, in the next stage, we advocate for a blind, multi-stage, and multi-annotator annotation process for the NCG scheme, recognizing it as a potentially better annotation model. We find that such a process while incorporating multiple worldviews could better address annotation inconsistencies that may have crept in in our current dataset. 

\paragraph{Non-uniform Distribution of Articles} As discussed earlier, our combined training dataset had 29 tasks and the test data had 10 tasks. However, these tasks did not have a uniform distribution of articles in our data. In the training data, the number of articles per task ranged from a maximum of 101 in one task, i.e. ``natural language inference,'' to a minimum of one article in seven tasks -  58.62\% of the training data tasks had less than 5 articles. The test dataset, on the other hand, followed a more uniform distribution than the training data ranging from a maximum of 32 articles to a minimum of seven articles at an average of 15.5\% articles per task. While our training dataset had over 200 articles, it may not have been sufficiently representative to learn uniform patterns. Thus in a new version of the dataset, a more uniform representation of the tasks will be attempted.

\section{Conclusions}

We have detailed the \textsc{NLPContributionGraph} Shared Task that entailed structuring research contributions in NLP articles as structured KGs. This task is the first-of-its-kind to be organized in the SemEval series. It attracted a strong participation demographic of 27 participants and seven teams - BERT transformer models were a popular choice among the participant systems in two different capacities, i.e. as classifiers or sequence labelers. Our task also saw the use of traditional parsers such a dependency syntax parsing technology. Further, some systems leveraged a hybrid approach including a combination of heuristics and machine learning. While the end-to-end task performance was low showing the task considerably challenging, each individual subtask toward obtaining an \textsc{NCG}, i.e. contribution sentence classification, scientific terms and relations extraction, and triples formation, demonstrated high performances in the subtask-only evaluation setting, i.e. when given gold data from the previous stage. The best system adopted a hybrid approach which seemed the most effective strategy for building the \textsc{NCG}. 

The \textsc{NCG} dataset is publicly available~\cite{ncg-final-dataset} and a KG overview of a structured form of our paper is here \url{https://www.orkg.org/orkg/comparison/R74774}.




\section*{Acknowledgments}
We thank the anonymous reviewers for their comments and suggestions. This work was co-funded by the European Research Council for the project ScienceGRAPH (Grant agreement ID: 819536) and by the TIB Leibniz Information Centre for Science and Technology.

\bibliographystyle{acl_natbib}
\bibliography{acl2021}

\begin{thebibliography}{31}
\expandafter\ifx\csname natexlab\endcsname\relax\def\natexlab#1{#1}\fi

\bibitem[{GRO(2008--2020)}]{GROBID}
 2008--2020.
\newblock \href
  {http://arxiv.org/abs/1:dir:6a298c1b2008913d62e01e5bc967510500f80710}
  {{GROBID}}.
\newblock \url{https://github.com/kermitt2/grobid}.

\bibitem[{Arora et~al.(2021)Arora, Kumar, Ghosal, Patwal, and
  Gooch}]{innovators}
Hardik Arora, Sandeep Kumar, Tirthankar Ghosal, Suraj Patwal, and Phil Gooch.
  2021.
\newblock {INNOVATORS at SemEval-2021 Task 11: A Dependency Parsing and
  BERT-based model for Extracting Contribution Knowledge from Scientific
  Papers}.
\newblock In \emph{Proceedings of the Fifteenth Workshop on Semantic
  Evaluation}, Bangkok (online). ACL.

\bibitem[{Auer et~al.(2020)Auer, Oelen, Haris, Stocker, D’Souza, Farfar,
  Vogt, Prinz, Wiens, and Jaradeh}]{auer2020improving}
S{\"o}ren Auer, Allard Oelen, Muhammad Haris, Markus Stocker, Jennifer
  D’Souza, Kheir~Eddine Farfar, Lars Vogt, Manuel Prinz, Vitalis Wiens, and
  Mohamad~Yaser Jaradeh. 2020.
\newblock {Improving Access to Scientific Literature with Knowledge Graphs}.
\newblock \emph{Bibliothek Forschung und Praxis}, 44(3):516--529.

\bibitem[{Auer(2018)}]{auer_soren_2018}
Sören Auer. 2018.
\newblock \href {https://doi.org/10.5281/zenodo.1157185} {{Towards an Open
  Research Knowledge Graph}}.

\bibitem[{Beltagy et~al.(2019)Beltagy, Lo, and Cohan}]{Beltagy2019SciBERT}
Iz~Beltagy, Kyle Lo, and Arman Cohan. 2019.
\newblock \href {http://arxiv.org/abs/arXiv:1903.10676} {{SciBERT: Pretrained
  Language Model for Scientific Text}}.
\newblock In \emph{EMNLP}.

\bibitem[{Birkle et~al.(2020)Birkle, Pendlebury, Schnell, and
  Adams}]{birkle2020web}
Caroline Birkle, David~A Pendlebury, Joshua Schnell, and Jonathan Adams. 2020.
\newblock Web of science as a data source for research on scientific and
  scholarly activity.
\newblock \emph{Quantitative Science Studies}, 1(1):363--376.

\bibitem[{Chen and Manning(2014)}]{chen2014fast}
Danqi Chen and Christopher~D Manning. 2014.
\newblock A fast and accurate dependency parser using neural networks.
\newblock In \emph{Proceedings of the 2014 conference on empirical methods in
  natural language processing (EMNLP)}, pages 740--750.

\bibitem[{Cho et~al.(2014)Cho, van Merri{\"e}nboer, Gulcehre, Bahdanau,
  Bougares, Schwenk, and Bengio}]{cho2014learning}
Kyunghyun Cho, Bart van Merri{\"e}nboer, Caglar Gulcehre, Dzmitry Bahdanau,
  Fethi Bougares, Holger Schwenk, and Yoshua Bengio. 2014.
\newblock {Learning Phrase Representations using RNN Encoder--Decoder for
  Statistical Machine Translation}.
\newblock In \emph{Proceedings of the 2014 Conference on Empirical Methods in
  Natural Language Processing (EMNLP)}, pages 1724--1734.

\bibitem[{Devlin et~al.(2019)Devlin, Chang, Lee, and
  Toutanova}]{devlin2019bert}
Jacob Devlin, Ming-Wei Chang, Kenton Lee, and Kristina Toutanova. 2019.
\newblock {BERT: Pre-training of Deep Bidirectional Transformers for Language
  Understanding}.
\newblock In \emph{Proceedings of the 2019 Conference of the North American
  Chapter of the Association for Computational Linguistics: Human Language
  Technologies, Volume 1 (Long and Short Papers)}, pages 4171--4186.

\bibitem[{D'Souza and Auer(2020)}]{ncg-pilot-scheme}
Jennifer D'Souza and S{\"o}ren Auer. 2020.
\newblock \href {http://ceur-ws.org/Vol-2658/paper2.pdf} {{NLPContributions: An
  Annotation Scheme for Machine Reading of Scholarly Contributions in Natural
  Language Processing Literature}}.
\newblock In \emph{Proceedings of the 1st Workshop on Extraction and Evaluation
  of Knowledge Entities from Scientific Documents (EEKE 2020) co-located with
  the ACM/IEEE Joint Conference on Digital Libraries in 2020 (JCDL 2020)},
  pages 16--27.

\bibitem[{D'Souza and Auer(2021)}]{d2020graphing}
Jennifer D'Souza and S{\"o}ren Auer. 2021.
\newblock \href {https://doi.org/doi:10.2478/jdis-2021-0023} {{Sentence,
  Phrase, and Triple Annotations to Build a Knowledge Graph of Natural Language
  Processing Contributions—A Trial Dataset}}.
\newblock \emph{Journal of Data and Information Science}, 0(0):--.

\bibitem[{D'Souza et~al.(2021)D'Souza, Auer, and Pederson}]{ncg-final-dataset}
Jennifer D'Souza, S{\"o}ren Auer, and Ted Pederson. 2021.
\newblock \href {https://zenodo.org/record/4737071} {{SemEval-2021 Task 11:
  NLPContributionGraph - Structuring Scholarly NLP Contributions for a Research
  Knowledge Graph}}.

\bibitem[{Fricke(2018)}]{fricke2018semantic}
Suzanne Fricke. 2018.
\newblock Semantic scholar.
\newblock \emph{Journal of the Medical Library Association: JMLA}, 106(1):145.

\bibitem[{He et~al.(2020)He, Liu, Gao, and Chen}]{he2020deberta}
Pengcheng He, Xiaodong Liu, Jianfeng Gao, and Weizhu Chen. 2020.
\newblock {DeBERTa}: Decoding-enhanced {BERT} with disentangled attention.
\newblock \emph{arXiv preprint arXiv:2006.03654}.

\bibitem[{Honnibal et~al.(2020)Honnibal, Montani, Van~Landeghem, and
  Boyd}]{spacy}
Matthew Honnibal, Ines Montani, Sofie Van~Landeghem, and Adriane Boyd. 2020.
\newblock \href {https://doi.org/10.5281/zenodo.1212303} {{spaCy:
  Industrial-strength Natural Language Processing in Python}}.

\bibitem[{Hu et~al.(2014)Hu, Lu, Li, and Chen}]{hu2014convolutional}
Baotian Hu, Zhengdong Lu, Hang Li, and Qingcai Chen. 2014.
\newblock {Convolutional Neural Network Architectures for Matching Natural
  Language Sentences}.
\newblock In \emph{NIPS}.

\bibitem[{Jaradeh et~al.(2019)Jaradeh, Oelen, Farfar, Prinz, D'Souza,
  Kismih{\'o}k, Stocker, and Auer}]{jaradeh2019open}
Mohamad~Yaser Jaradeh, Allard Oelen, Kheir~Eddine Farfar, Manuel Prinz,
  Jennifer D'Souza, G{\'a}bor Kismih{\'o}k, Markus Stocker, and S{\"o}ren Auer.
  2019.
\newblock Open research knowledge graph: next generation infrastructure for
  semantic scholarly knowledge.
\newblock In \emph{Proceedings of the 10th International Conference on
  Knowledge Capture}, pages 243--246.

\bibitem[{Johnson et~al.(2018)Johnson, Watkinson, and Mabe}]{johnson2018stm}
Rob Johnson, Anthony Watkinson, and Michael Mabe. 2018.
\newblock \href {https://www.stm-assoc.org/2018_10_04_STM_Report_2018.pdf}
  {{The STM report}}.
\newblock \emph{An overview of scientific and scholarly publishing. 5th edition
  October}.

\bibitem[{Landhuis(2016)}]{landhuis2016scientific}
Esther Landhuis. 2016.
\newblock Scientific literature: information overload.
\newblock \emph{Nature}, 535(7612):457--458.

\bibitem[{Lin et~al.(2021)Lin, Liu, Ling, Wang, Chen, and He}]{ecnuica}
Jiaju Lin, Jiawei Liu, Jing Ling, Zhiwei Wang, Qin Chen, and Liang He. 2021.
\newblock {ECNUICA at SemEval-2021 Task 11: Schema based Information Extraction
  Pipeline}.
\newblock In \emph{Proceedings of the Fifteenth Workshop on Semantic
  Evaluation}, Bangkok (online). ACL.

\bibitem[{Liu et~al.(2021)Liu, Sarol, and Kilicoglu}]{uiuc-bionlp}
Haoyang Liu, Janina Sarol, and Halil Kilicoglu. 2021.
\newblock {UIUC\_BioNLP at SemEval-2021 Task 11: A Cascade of Neural Models for
  Structuring Scholarly NLP Contributions}.
\newblock In \emph{Proceedings of the Fifteenth Workshop on Semantic
  Evaluation}, Bangkok (online). ACL.

\bibitem[{Liu et~al.(2019)Liu, Ott, Goyal, Du, Joshi, Chen, Levy, Lewis,
  Zettlemoyer, and Stoyanov}]{liu2019roberta}
Yinhan Liu, Myle Ott, Naman Goyal, Jingfei Du, Mandar Joshi, Danqi Chen, Omer
  Levy, Mike Lewis, Luke Zettlemoyer, and Veselin Stoyanov. 2019.
\newblock {RoBERTa}: A robustly optimized {BERT} pretraining approach.
\newblock \emph{arXiv preprint arXiv:1907.11692}.

\bibitem[{Ma et~al.(2021)Ma, Wang, and Zhang}]{ynu-hpcc}
Xinge Ma, Jin Wang, and Xuejie Zhang. 2021.
\newblock {YNU-HPCC at SemEval-2021 Task 11: Using a BERT Model to Extract
  Contributions from NLP Scholarly Articles}.
\newblock In \emph{Proceedings of the Fifteenth Workshop on Semantic
  Evaluation}, Bangkok (online). ACL.

\bibitem[{Manghi et~al.(2019)Manghi, Atzori, Bardi, Shirrwagen, Dimitropoulos,
  La~Bruzzo, Foufoulas, Löhden, Bäcker, Mannocci, Horst, Baglioni, Czerniak,
  Kiatropoulou, Kokogiannaki, De~Bonis, Artini, Ottonello, Lempesis, Nielsen,
  Ioannidis, Bigarella, and Summan}]{manghi_paolo_2019_3516918}
Paolo Manghi, Claudio Atzori, Alessia Bardi, Jochen Shirrwagen, Harry
  Dimitropoulos, Sandro La~Bruzzo, Ioannis Foufoulas, Aenne Löhden, Amelie
  Bäcker, Andrea Mannocci, Marek Horst, Miriam Baglioni, Andreas Czerniak,
  Katerina Kiatropoulou, Argiro Kokogiannaki, Michele De~Bonis, Michele Artini,
  Enrico Ottonello, Antonis Lempesis, Lars~Holm Nielsen, Alexandros Ioannidis,
  Chiara Bigarella, and Friedrich Summan. 2019.
\newblock \href {https://doi.org/10.5281/zenodo.3516918} {{OpenAIRE Research
  Graph Dump}}.

\bibitem[{Martin and Pedersen(2021)}]{duluth}
Anna Martin and Ted Pedersen. 2021.
\newblock {Duluth at SemEval-2021 Task 11: Applying DeBERTa to Contributing
  Sentence Selection and Dependency Parsing for Entity Extraction}.
\newblock In \emph{Proceedings of the Fifteenth Workshop on Semantic
  Evaluation}, Bangkok (online). ACL.

\bibitem[{Qi et~al.(2020)Qi, Zhang, Zhang, Bolton, and Manning}]{qi2020stanza}
Peng Qi, Yuhao Zhang, Yuhui Zhang, Jason Bolton, and Christopher~D. Manning.
  2020.
\newblock \href {https://nlp.stanford.edu/pubs/qi2020stanza.pdf} {Stanza: A
  {Python} natural language processing toolkit for many human languages}.
\newblock In \emph{Proceedings of the 58th Annual Meeting of the Association
  for Computational Linguistics: System Demonstrations}.

\bibitem[{Shailabh et~al.(2021)Shailabh, Chaurasia, and Modi}]{knowgraph-iitk}
Shashank Shailabh, Sajal Chaurasia, and Ashutosh Modi. 2021.
\newblock {KnowGraph@IITK at SemEval-2021 Task 11: Building Knowledge Graph for
  NLP Research}.
\newblock In \emph{Proceedings of the Fifteenth Workshop on Semantic
  Evaluation}, Bangkok (online). ACL.

\bibitem[{Souza et~al.(2019)Souza, Nogueira, and Lotufo}]{souza2019portuguese}
F{\'a}bio Souza, Rodrigo Nogueira, and Roberto Lotufo. 2019.
\newblock Portuguese named entity recognition using {BERT-CRF}.
\newblock \emph{arXiv preprint arXiv:1909.10649}.

\bibitem[{Wang et~al.(2020)Wang, Shen, Huang, Wu, Dong, and
  Kanakia}]{wang2020microsoft}
Kuansan Wang, Zhihong Shen, Chiyuan Huang, Chieh-Han Wu, Yuxiao Dong, and
  Anshul Kanakia. 2020.
\newblock {Microsoft academic graph: When experts are not enough}.
\newblock \emph{Quantitative Science Studies}, 1(1):396--413.

\bibitem[{Zhang et~al.(2021)Zhang, Su, He, Lin, Sun, and Shan}]{itnlp}
Genyu Zhang, Yu~Su, Changhong He, Lei Lin, Chengjie Sun, and Lili Shan. 2021.
\newblock {ITNLP at SemEval-2021 Task 11: Boosting BERT with Sampling and
  Adversarial Training for Knowledge Extraction}.
\newblock In \emph{Proceedings of the Fifteenth Workshop on Semantic
  Evaluation}, Bangkok (online). ACL.

\bibitem[{Zhang et~al.(2020)Zhang, Xu, Han, Niu, Cui, Sugiyama, and
  Kankanhalli}]{zhang2020attacks}
Jingfeng Zhang, Xilie Xu, Bo~Han, Gang Niu, Lizhen Cui, Masashi Sugiyama, and
  Mohan Kankanhalli. 2020.
\newblock Attacks which do not kill training make adversarial learning
  stronger.
\newblock In \emph{International Conference on Machine Learning}, pages
  11278--11287. PMLR.

\end{thebibliography}

\appendix

\section{Per Information Unit Evaluations}

Table 7 shows triple extraction F1 scores for each of the IUs. The scores from each of the three evaluation phases in our Shared Task are separated by slash symbols. Recall that from the second evaluation phase, the gold data were made available to the systems starting with sentences (\textit{Evaluation Phase 2, Part 1: Phrases and Triples}) followed by the terms and relations additionally (\textit{Evaluation Phase 2, Part 2: Triples}). 

Comparing the performances across IUs, we see the \textsc{Code} IU was the easiest to extract. In \textit{Phase 1}, the best F1 was 83.33\%. In both \textit{Phase 2, Part 1} and \textit{Part 2}, the best F1 was 100.0\%. This is an expected result for \textsc{Code} to be easiest to extract since it had the simplest annotation patterns; an example is depicted in Fig. 5. 

\begin{table}[!htb]
\small
\resizebox{\linewidth}{!}{\begin{tabular}{ll|r|r}
&                          & Training Data & Test Data \\ \cline{2-4}
1 & \textsc{Code}              & 1.0           & 1.03      \\
2 & \textsc{ResearchProblem}   & 2.78          & 2.13      \\
3 & \textsc{Dataset}           & 13.78         & 22.5      \\
4 & \textsc{Approach}          & 15.55         & 17.61     \\
5 & \textsc{Model}             & 18.06         & 20.14     \\
6 & \textsc{AblationAnalysis}  & 18.82         & 21.19     \\
7 & \textsc{Baselines}         & 20.13         & 15.3      \\
8 & \textsc{Results}           & 23.31         & 23.0      \\
9 & \textsc{Hyperparameters}   & 23.78         & 19.78     \\
10 & \textsc{ExperimentalSetup} & 27.53         & 27.47     \\
11 & \textsc{Tasks}             & 34.63         & -         \\
12 & \textsc{Experiments}       & 54.65         & 39.06    
\end{tabular}}
\caption{Average no. of triples per Information Unit}
\end{table}

Table 6 shows the average number of triples per IU reflecting, in a sense, their complexity. We hypothesize that the more the triples, the more complex the extraction task. Comparing these numbers with the results in Table 7, we see that 5th ranked IU, i.e. \textsc{Model}, showed the next easiest to extract after \textsc{Code}, at 38.14\% F1, in the end-to-end setting. Following which, we see that the 2nd ranked IU, i.e. \textsc{ResearchProblem}, obtained an F1 of 35.79\%. Nevertheless, confirming our hypothesis, we found a negative correlation (\textit{r} -0.65) between the training data triples size per IU and the end-to-end system performances, i.e. for IUs with fewer triples the extraction score is higher for most IUs. The negative correlations were progressively stronger from \textit{Part 1} to \textit{Part 2} in \textit{Evaluation Phase 2} (\textit{r} -0.75 and \textit{r} -0.79), respectively.

\begin{figure}[!htb]
  \includegraphics[width=\linewidth]{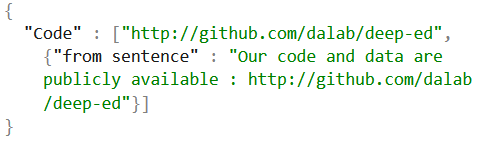}
  \caption{Annotated data in JSON format for the \textsc{Code} Information Unit for the paper ``Deep Joint Entity Disambiguation with Local Neural Attention.''}  
\end{figure}

\begin{table*}[!htb]
    \begin{subtable}{\textwidth}
    \small
\begin{tabular}{|l|r|r|r|r|} \hline
\textbf{Team}  & \multicolumn{1}{c|}{\underline{\textsc{ResearchProblem}}} & \multicolumn{1}{c|}{\underline{\textsc{Approach}}} & \multicolumn{1}{c|}{\underline{\textsc{Model}}} & \multicolumn{1}{c|}{\underline{\textsc{Code}}} \\
UIUC\_BioNLP   & 26.17/ 53.19/ 89.41         & \textbf{11.54}/ \textbf{20.2}/ 28.87       & \textbf{38.14}/ \textbf{55.31}/ \textbf{76.9}        & 57.14/ 80.0/ \textbf{100.0} \\
ITNLP & \textbf{35.79}/ 43.18/ 78.35         & 0.0/ 0.0/ 0.0            & 16.03/ 22.65/ 51.42       & \textbf{83.33}/ \textbf{100.0}/ \textbf{100.0} \\
KnowGraph@IITK & 24.62/ 25.81/ \textbf{97.56}         & 4.94/ 5.93/ 0.0          & 8.2/ 19.18/ 34.48         & \textbf{83.33}/ \textbf{100.0}/ \textbf{100.0} \\
ECNUICA        & 6.45/ \textbf{65.12}/ 89.89          & 1.75/ 17.83/ \textbf{28.93}       & 0.0/ 29.73/ 56.67         & 0.0/ 80.0/ 80.0 \\
INNOVATORS     & 9.88/ 9.88/ 3.25            & 0.0/ 0.0/ 3.8            & 0.0/ 0.0/ 7.55            & 50.0/ 50.0/ 0.0 \\
DULUTH         & 0.0/ 4.71/ 58.73            & 0.0/ 2.06/ 21.78         & 7.23/ 7.14/ 35.36         & 0.0/ 40.0/ 88.89 \\
YNU-HPCC       & -/ 2.9/ 5.07                & -/ 2.53/ 7.22            & -/ 3.52/ 18.29            & -/ 0.93/ 0.56 \\ \hline           
\end{tabular}
    \end{subtable}
    \vfill
    \begin{subtable}{\textwidth}
    \small
\begin{tabular}{|l|r|r|r|} \hline
\textbf{Team}  & \multicolumn{1}{c|}{\underline{\textsc{ExperimentalSetup}}} & \multicolumn{1}{c|}{\underline{\textsc{Hyperparameters}}} & \multicolumn{1}{c|}{\underline{\textsc{Baselines}}} \\
UIUC\_BioNLP & \textbf{28.37}/ \textbf{52.42}/ \textbf{67.27} & \textbf{5.6}/ \textbf{35.71}/ \textbf{39.44} & \textbf{20.69}/ \textbf{50.85}/ \textbf{74.34} \\
ITNLP   & 18.78/ 29.87/ 42.16  & 4.0/ 6.06/ 12.63  & 0.0/ 16.67/ 27.69 \\
KnowGraph@IITK & 12.14/ 13.53/ 12.7  & 4.37/ 7.88/ 8.48     & 3.47/ 6.78/ 33.33 \\
ECNUICA & 14.79/ 25.88/ 42.34  & 3.36/ 13.4/ 3.36   & 9.11/ 34.62/ 51.06 \\
INNOVATORS     & 0.0/ 0.0/ 0.0   & 0.0/ 0.0/ 0.0    & 0.0/ 0.0/ 0.0 \\
DULUTH         & 1.54/ 6.33/ 30.47  & 4.04/ 7.89/ 21.43  & 5.56/ 3.23/ 17.5 \\
YNU-HPCC       & -/ 4.24/ 16.7     & -/ 0.88/ 2.58      & -/ 0.87/ 3.5 \\ \hline           
\end{tabular}
    \end{subtable}
    \vfill    
    \begin{subtable}{\textwidth}
    \small
\begin{tabular}{|l|r|r|r|} \hline
\textbf{Team} & \multicolumn{1}{c|}{\underline{\textsc{Results}}} & \multicolumn{1}{c|}{\underline{\textsc{Experiments}}} & \multicolumn{1}{c|}{\underline{\textsc{AblationAnalysis}}} \\
UIUC\_BioNLP  & \textbf{20.62}/ \textbf{37.77}/ \textbf{56.4}  & 7.19/ \textbf{8.96}/ 10.61       & \textbf{23.01}/ \textbf{31.78}/ \textbf{61.36}     \\
ITNLP  & 8.85/ 17.47/ 42.5       & 1.48/ 1.42/ 0.0         & 8.6/ 6.35/ 11.63            \\
KnowGraph@IITK & 10.86/ 17.55/ 28.94     & 3.33/ 1.96/ 0.0         & 4.23/ 3.74/ 35.16           \\
ECNUICA  & 15.37/ 26.25/ 49.0      & \textbf{7.86}/ 6.06/ \textbf{13.86}       & 3.94/ 4.6/ 8.82             \\
INNOVATORS & 0.0/ 0.0/ 7.36 & 0.0/ 0.0/ 0.0           & 0.0/ 0.0/ 0.0               \\
DULUTH & 7.2/ 8.04/ 30.96   & 0.0/ 1.72/ 5.97         & 0.0/ 0.0/ 12.29             \\
YNU-HPCC  & -/ 3.56/ 16.63          & -/ 0.68/ 2.73           & -/ 1.05/ 2.79 \\ \hline        
\end{tabular}
\end{subtable}
     \caption{Per Information Unit F1 scores per evaluation phase of the seven participating teams. The three scores in each row are from the three evaluation phases in the Shared Task as follows [\textit{Evaluation Phase 1: End-to-end Pipeline}]\textbf{/}[\textit{Evaluation Phase 2, Part 1: Phrases and Triples}]\textbf{/}[\textit{Evaluation Phase 2, Part 2: Triples}]. \footnotesize{Best scores are in bold.}}
\end{table*}

\begin{table*}[!htb]
\small
\centering
\begin{tabular}{|l|l|l|l|c|l|l|} \hline
\multicolumn{1}{|c|}{\multirow{2}{*}{Model}} & \multicolumn{3}{c|}{Information Units} & \multicolumn{3}{c|}{Triples}            \\ \cline{2-7} 
\multicolumn{1}{|c|}{} & F1 & P & R & \multicolumn{1}{l|}{F1} & P & R     \\ \hline
UIUC\_BioNLP    & \textbf{72.93}/\textbf{83.98} & 66.67/76.77 & 80.49/92.68 & \textbf{22.28}/\textbf{25.01} & 22.3/25.08  & 22.26/24.94 \\ \hline
ITNLP & \textbf{72.93}/82.49 & 66.67/76.84 & 80.49/89.02 & 13.79/14.26 & 13.39/13.98 & 14.23/14.56 \\ \hline
KnowGraph@IITK  & 60.54/72.32 & 44.13/57.04 & 96.34/98.78 & 8.57/10.0 & 6.53/7.87 & 12.45/13.72 \\ \hline
ECNUICA & 54.05/56.76 & 42.86/45.0 & 73.17/76.83 & 6.78/6.72 & 4.28/4.24 & 16.29/16.12 \\ \hline
INNOVATORS & 71.72/80.0 & 82.54/92.06 & 63.41/70.73 & 0.97/0.97  & 14.29/14.29 & 0.5/0.5   \\ \hline
DULUTH & 64.41/77.11 & 60.0/76.19 & 69.51/78.05 & 3.94/4.05  & 9.2/10.42 & 2.51/2.51  \\ \hline
\end{tabular}
\caption{\textit{Evaluation Phase 1: End-to-end Pipeline} Results with (\textsc{Approach}, \textsc{Model}) IUs normalized to \textsc{Approach} and (\textsc{ExperimentalSetup}, \textsc{Hyperparameters}) IUs normalized to \textsc{ExperimentalSetup}. \footnotesize{Best scores are in bold. Scores before the slash are from original dataset and scores after the slash are from the normalized dataset.}}
\end{table*}

\section{Normalized \textsc{Approach} and \textsc{ExperimentalSetup} Evaluations}

In Table 8, we revisit overall scores from Table 5 for two evaluation aspects in the end-to-end system evaluations, i.e. only for extracting Information Units and Triples. We revisit just these two aspects because they were impacted when we obtained normalizations of four IU labels into two, respectively, i.e. \textsc{Approach} and \textsc{Model} as \textsc{Approach} and \textsc{ExperimentalSetup} and \textsc{Hyperparameters} as \textsc{ExperimentalSetup}. By this, we can observe system performances on a simplified version of our task. Observing ``Triples'' F1, we see that the ordering of the system performance without and with normalization remain unchanged - the best score obtained a 3 points boost.

\end{document}